\documentclass[11pt]{article}
\usepackage[preprint]{acl}
\usepackage{times}
\usepackage{latexsym}
\usepackage{xspace}
\usepackage[T1]{fontenc}
\usepackage[utf8]{inputenc}
\usepackage{microtype}
\usepackage{inconsolata}
\usepackage{graphicx}

\usepackage{booktabs}
\usepackage[framemethod=TikZ]{mdframed}
\usepackage{tcolorbox}
\tcbuselibrary{breakable}
\usepackage{enumitem}
\usepackage{multirow}
\usepackage{bigstrut}
\usepackage[ruled,vlined]{algorithm2e}
\usepackage[accsupp]{axessibility}

\tcbuselibrary{skins}
\usepackage{booktabs}
\usepackage{tabularx}
\usepackage{rotating}

\title{Understanding Machine Taste: Aesthetic Agreement Across Foundation Models in K-pop}

\author{
Kaishen Wang,
Yucheng Pan,
Xuehao Cui,
Hanling Tian,
Heng Huang \\
University of Maryland, College Park \\
\texttt{\{kaishen, ycpan, cedrus, hanling, heng\}@umd.edu}
}

\begin{document}
\maketitle

\begin{abstract}
As foundation models enter creative industries, understanding their aesthetic preferences becomes an important step toward AI-assisted creative decision making. In this work, we investigate how foundation models evaluate creative content and characterize their ``machine taste'' using K-pop songs and music videos as a testbed. We formulate aesthetic evaluation as a pairwise ranking problem and systematically examine pairwise inter-model agreement, group-level consensus, and cross-modal preference alignment. Our results reveal stronger pairwise inter-model agreement and group-level consensus in visual evaluation than in audio evaluation, while preferences across the two modalities remain weakly aligned. We further examine how visual consensus rankings relate to real-world popularity using historical YouTube viewership data, finding positive but varying correlations across artist groups. These findings provide initial insights into machine aesthetic preferences and lay the groundwork for future AI-assisted creative decision making that considers creative content, artist identity, audience preferences, and market dynamics.
\end{abstract}

\section{Introduction}

Recent foundation models~\cite{ma2024foundation,yang2025qwen3,xu2025qwen3,yang2026moss,ghosh2026music,ye2026videoresearcher,wang2026look} have become increasingly capable across language, vision, and audio, and are gradually being integrated into real-world workflows beyond conventional benchmark-oriented tasks. This transition is particularly important in domains where decisions are not determined by a single correct answer. In creative industries, for example, choices such as selecting a song, developing a visual concept, or evaluating creative content are inherently subjective and often depend on aesthetic judgment rather than objective correctness~\cite{zhao2025language,fein2026litbench,chen2026llms}.

The entertainment industry provides a timely example of this transition, as major companies are increasingly exploring AI applications in their creative and production workflows. SM Entertainment~\cite{sm2026ceomessage} has identified AI as a key component of its SM NEXT 3.0 strategy, aiming to improve production efficiency and support more sophisticated fan- and market-informed decision making. HYBE~\cite{hybe2026technology} has expanded its AI capabilities through Supertone, which develops AI-based audio technologies for generating realistic singing and acting voices. JYP Entertainment~\cite{jyp2025q1} has similarly expanded its AI-related initiatives, including plans to develop an AI artist as part of its future content portfolio. These developments reflect the growing role of AI in creative industries and point toward broader opportunities for AI-assisted creative and strategic decision making. An important first step toward this vision is to understand how foundation models themselves evaluate creative content and what aesthetic preferences they exhibit.

\begin{figure*}
    \centering
    \includegraphics[width=0.95\linewidth]{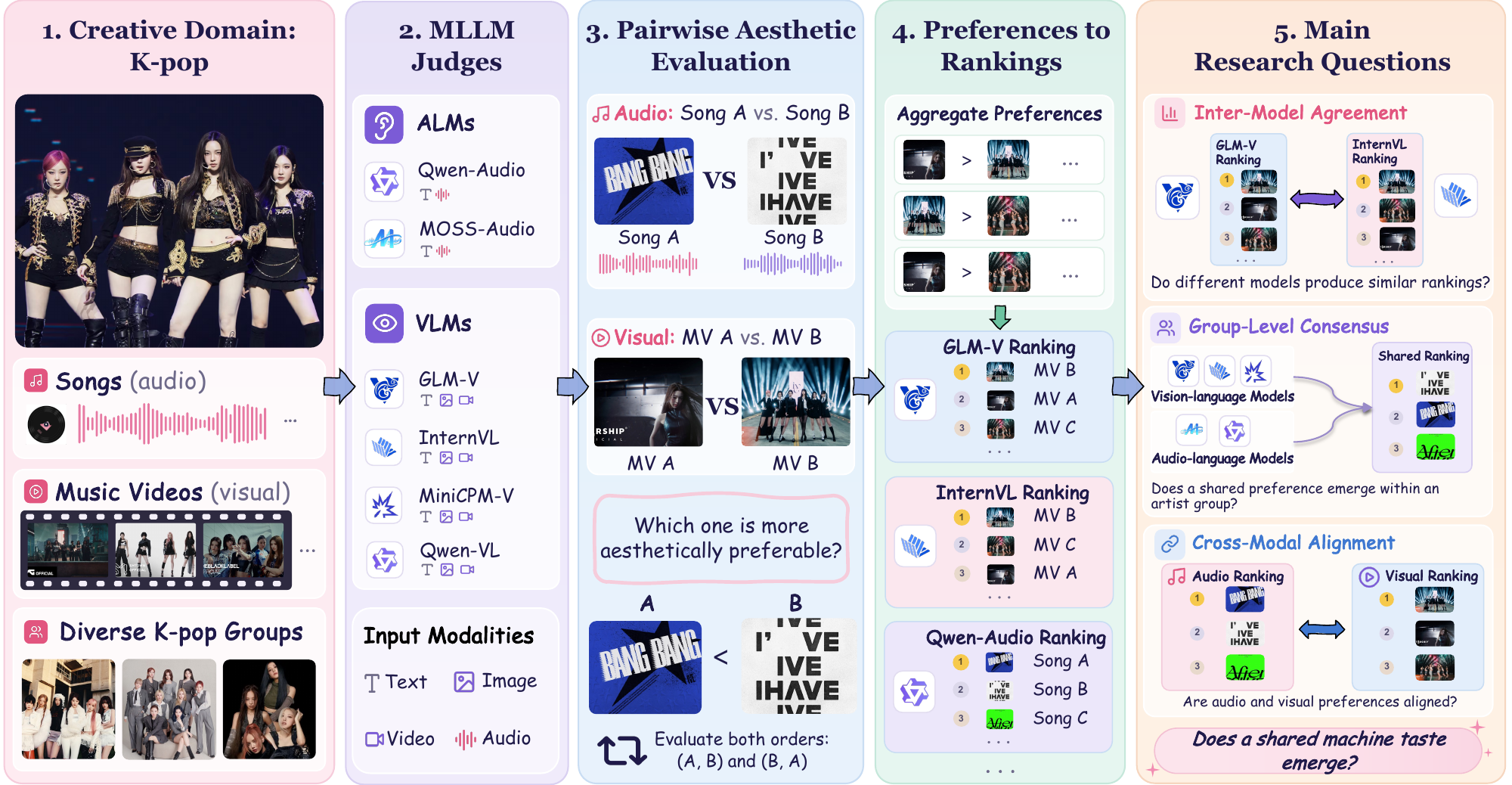}
    \caption{Overview of our study of machine taste in K-pop. ALMs and VLMs perform pairwise aesthetic evaluations of songs and music videos, respectively. The resulting rankings are analyzed to examine inter-model agreement, group-level consensus, and cross-modal preference alignment.}
    \vspace{-2mm}
\label{fig:overview}

\end{figure*}

Unlike factual question answering or mathematical reasoning, aesthetic evaluation does not admit a unique ground-truth answer~\cite{marco2025reader}. While prior studies have investigated the aesthetic evaluation capabilities and biases of foundation models~\cite{li2024dissecting,yao2025multimodal,li2025aesbiasbench}, understanding how these models evaluate creative content remains challenging without a universally accepted reference for aesthetic quality. In this context, examining aesthetic preference agreement provides a useful starting point for investigating machine taste, allowing us to examine whether shared preferences emerge between individual models, among models evaluating the same artist group, and across evaluation modalities.

To investigate machine taste through aesthetic preference agreement, we use K-pop as a testbed that offers diverse creative content across both audio and visual media. Songs and music videos span a wide range of musical styles, production choices, and visual concepts, providing rich material for examining aesthetic preferences. Moreover, the availability of audio tracks and corresponding music videos enables systematic comparisons of model preferences within and across modalities.

% \vspace{-0.4mm}
In this work, we address three research questions, as illustrated in Figure~\ref{fig:overview}: (1) how strongly individual models agree in their aesthetic rankings across all artist groups, (2) whether a broader consensus emerges among all evaluated models within each artist group, and (3) whether aesthetic preferences align across audio and visual modalities. To investigate these questions, we formulate aesthetic evaluation as a pairwise ranking problem, considering both audio-based song ranking and visual-only music-video ranking. We analyze the resulting preferences using Spearman rank correlation and Kendall's coefficient of concordance. Our results show stronger agreement between individual vision-language models (VLMs) and consistently higher group-level consensus in visual evaluation across all 14 artist groups, compared with audio evaluation by audio-language models (ALMs). Meanwhile, audio and visual preferences remain weakly aligned. We also examine how model-derived visual preferences relate to real-world popularity using historical YouTube viewership data, finding positive but varying correlations across artist groups.

More broadly, this study represents a first step toward AI-assisted decision making in the K-pop industry. Understanding machine aesthetic preferences can help inform future systems that jointly consider creative content, artist identity, audience preferences, and market dynamics to support broader creative and strategic decisions. Our contributions are summarized as follows:

\vspace{-1mm}

\begin{itemize}
    \item We introduce a systematic evaluation framework for investigating machine aesthetic preferences through pairwise ranking, inter-model agreement, group-level consensus, and cross-modal preference alignment.

    \vspace{-1mm}

    \item We conduct a large-scale empirical study of foundation models using K-pop songs and music videos, revealing weaker and more variable aesthetic agreement in audio evaluation, stronger and more consistent agreement in visual evaluation, and limited alignment between the two modalities.

    \vspace{-1mm}

    \item We further examine the relationship between model-derived visual preferences and real-world popularity using historical YouTube viewership data, finding positive but varying correlations across artist groups.
\end{itemize}

\section{Related Work}

\subsection{Multimodal Models for Creative Content}
Recent foundation models have extended multimodal understanding from text and images to audio and video, enabling direct perception and reasoning over increasingly complex forms of creative content~\cite{ma2024foundation,xu2025qwen3,yang2026moss,ghosh2026music}. Audio-language models (ALMs) such as Qwen3-Omni~\cite{xu2025qwen3} and MOSS~\cite{yang2026moss} can directly process acoustic signals, including music, while recent vision-language models (VLMs) such as Qwen3-VL~\cite{bai2025qwen3}, InternVL3.5~\cite{wang2025internvl3}, GLM-4.6V-Flash~\cite{hong2025glm}, MiniCPM-V-4.5~\cite{yu2026minicpm}, and Qwen3.5~\cite{qwen3.5}, provide strong capabilities for understanding videos. Existing evaluations of these models primarily focus on perceptual understanding, grounding, question answering, and multimodal reasoning. In contrast, our work studies a different aspect of multimodal intelligence: whether independently developed foundation models form similar subjective preferences when evaluating the same creative content.

\subsection{Aesthetic Evaluation and Model Agreement}

Recent studies have investigated the ability of foundation models to evaluate subjective creative content, including visual aesthetic sensitivity, alignment with human aesthetic preferences, and the evaluation of creative writing~\cite{yao2025multimodal,li2025aesbiasbench,feng2026visual,kim2025evaluating,jiang2026artificial}. These studies suggest that foundation models can capture non-trivial aesthetic signals, but their judgments remain imperfect and may differ substantially from human or expert preferences. Other studies have examined the reliability of LLMs and MLLMs as evaluators~\cite{chen2024humans,ye2025justice,shi2025judging,feng2025we}, identifying systematic biases such as position bias, authority and presentation effects, and model-specific preferences. These studies also highlight inconsistencies in model judgments when evaluating the same content. Other studies have explored ways to improve the consistency of model-based evaluation through stronger reasoning or specialized evaluation models~\cite{saha2025learning,xiong2025lllava,xiong2026multi}. These efforts primarily focus on making model judgments more reliable and better aligned with human preferences, rather than understanding the disagreement itself. In contrast, our work focuses on inter-model aesthetic agreement rather than evaluation accuracy. We examine whether independently developed foundation models share similar aesthetic preferences and whether these preferences remain consistent across audio and visual modalities.
\section{Method}
\label{sec:method}

Our goal is to investigate how foundation models evaluate creative content through their expressed aesthetic preferences. 
Rather than relying on absolute quality scores, which may provide limited discrimination when models assign similar scores to different creative items, we formulate aesthetic evaluation as a pairwise ranking problem to directly capture models' relative preferences. Models compare candidate items, and their pairwise preferences are aggregated into model-specific rankings. We then examine how consistently models rank creative content within each modality and whether their preferences align across audio and visual evaluation.

% Our goal is to study whether independently developed foundation models exhibit consistent aesthetic preferences over creative content. Rather than relying on absolute quality scores, which may not be directly comparable because different models may use different scoring standards, we formulate aesthetic evaluation as a pairwise ranking problem. Models directly compare candidate items, and their pairwise preferences are aggregated into model-specific rankings. We then analyze the agreement among these rankings within each modality and examine whether preference structures remain aligned across audio and visual evaluation.

\subsection{Pairwise Aesthetic Ranking}
\label{sec:pairwise_ranking}

Consider an evaluation set
\begin{equation}
\mathcal{C}=\{c_1,c_2,\ldots,c_N\},
\end{equation}
where each $c_i$ denotes a creative item to be evaluated. Depending on the evaluation setting, a candidate corresponds to either an audio track or a music video.

For each unordered pair $\{c_i,c_j\}$, we evaluate model $m$ under both presentation orders:
\begin{equation}
(c_i,c_j)
\qquad \text{and} \qquad
(c_j,c_i).
\end{equation}
The two evaluations are treated as independent comparisons. Each comparison requires a binary preference decision, and the model must select exactly one of the two candidates. The preferred candidate receives one point. Evaluating both presentation orders reduces the influence of positional preference while retaining the model's individual decision in each comparison.

Let
\begin{equation}
y_m(c_i,c_j)\in\{0,1\}
\end{equation}
denote the model judgment when $c_i$ is presented before $c_j$, where $y_m(c_i,c_j)=1$ indicates that $c_i$ is preferred over $c_j$. The total score assigned to candidate $c_i$ is
\begin{equation}
s_m(c_i)
=
\sum_{j\neq i}
\left[
y_m(c_i,c_j)
+
1-y_m(c_j,c_i)
\right].
\label{eq:aggregate_score}
\end{equation}

The final ranking produced by model $m$ is obtained by sorting all candidates according to their aggregate preference scores:
\begin{equation}
R_m
=
\operatorname{Rank}
\left(
s_m(c_1),
s_m(c_2),
\ldots,
s_m(c_N)
\right).
\label{eq:model_ranking}
\end{equation}

This procedure provides a common ranking framework across models while avoiding direct comparison of model-specific absolute scoring scales.

\subsection{Aesthetic Preference Agreement}
\label{sec:agreement}

Given the rankings produced by different models, we analyze aesthetic agreement from two complementary perspectives: how consistently individual models rank creative content across all artist groups, and whether a broader consensus emerges among models within each group.

\paragraph{Pairwise Inter-model Agreement.}
For two models $m_i$ and $m_j$, we measure the similarity between their induced rankings using Spearman's rank correlation:
\begin{equation}
\rho_{i,j}^{(g)}
=
\rho_{\mathrm{S}}
\left(
R_{m_i,g},
R_{m_j,g}
\right),
\label{eq:spearman}
\end{equation}
where $g$ indexes an artist group. Given $G$ artist groups, we report the mean pairwise agreement across groups:
\begin{equation}
\bar{\rho}_{i,j}
=
\frac{1}{G}
\sum_{g=1}^{G}
\rho_{i,j}^{(g)}.
\label{eq:avg_spearman}
\end{equation}
This metric characterizes how similarly a particular pair of models ranks creative content across different artist groups.

\paragraph{Group-level Model Consensus.}
While pairwise correlation compares individual model pairs, we also examine whether all evaluated models exhibit a shared preference ranking within each artist group. For each group $g$, we therefore compute Kendall's coefficient of concordance over the rankings produced by all $M$ models:
\begin{equation}
W_g
=
W
\left(
R_{m_1,g},
R_{m_2,g},
\ldots,
R_{m_M,g}
\right),
\label{eq:kendall_w}
\end{equation}
where $W_g\in[0,1]$. A larger value indicates stronger collective agreement among models, with $W_g=1$ corresponding to identical rankings.

% The two metrics capture complementary aspects of aesthetic agreement: $\bar{\rho}_{i,j}$ measures how similarly a particular pair of models behaves across artist groups, whereas $W_g$ measures how strongly the full model population agrees within a particular group.

\begin{table}[t]
\centering
\small
\begin{tabular}{l cc}
\toprule
\textbf{Group} & \textbf{\# Audio} & \textbf{\# Video} \\
\midrule
RED VELVET     & 30 & 28 \\
TWICE          & 30 & 30 \\
BLACKPINK      & 30 & 30 \\
I-DLE          & 30 & 25 \\
ITZY           & 30 & 26 \\
AESPA          & 30 & 23 \\
IVE            & 30 & 23 \\
NMIXX          & 30 & 16 \\
LE SSERAFIM    & 30 & 17 \\
NEWJEANS       & 20 & 16 \\
BABYMONSTER    & 24 & 16 \\
ILLIT          & 30 & 12 \\
MEOVV          & 13 & 7 \\
HEARTS2HEARTS  & 16 & 9 \\
\midrule
\textbf{Total} & \textbf{373} & \textbf{278} \\
\bottomrule
\end{tabular}
\caption{
Statistics of the K-pop evaluation dataset across 14 artist groups. The audio collection includes up to 30 songs per group, while the video collection contains the subset with eligible official music videos.
}
\vspace{-4mm}
\label{tab:dataset_statistics}
\end{table}

\subsection{Modality-specific and Cross-modal Preference Analysis}
\label{sec:evaluation_settings}

We apply the pairwise ranking framework in Section~\ref{sec:pairwise_ranking} to audio-based song evaluation and music-video-based evaluation, both conducted within individual artist groups. The resulting rankings are analyzed using the agreement metrics defined in Section~\ref{sec:agreement}. We further examine preference alignment between audio and visual evaluation. In all evaluations, models are not provided with song titles, popularity statistics, chart performance, or other external market information. Detailed prompts are presented in Appendix~\ref{app:prompts}.

\paragraph{Audio-based Preference Analysis.}
For each K-pop artist group $g$, we construct a song set $\mathcal{S}_g=\{s_1,\ldots,s_{N_g}\}$ and evaluate audio-language models (ALMs) using only the audio content. Models compare songs based on overall musical quality, including melody, vocal performance, production, arrangement, and overall cohesion. We obtain the ranking $R^{\mathrm{audio}}_{m,g}$ for each ALM $m$ following Section~\ref{sec:pairwise_ranking}, and analyze agreement among ALMs using the metrics in Section~\ref{sec:agreement}.

\paragraph{Visual Preference Analysis.}
For each artist group $g$, we construct a set of official music videos $\mathcal{V}_g=\{v_1,\ldots,v_{M_g}\}$. Vision-language models (VLMs) receive only the visual stream of each music video and are not given access to the corresponding audio track. Models compare videos based on visual and creative execution, including visual concept, cinematography, styling, editing, and overall visual coherence. Following the same procedure, we obtain the ranking $R^{\mathrm{MV}}_{m,g}$ for each VLM $m$ and analyze agreement among VLMs.

\paragraph{Cross-modal Preference Alignment.}
We examine whether audio and visual aesthetic preferences align for the same underlying songs. For each artist group $g$, we restrict the analysis to songs shared by the audio and music-video evaluation sets. We recompute the ALM rankings using only pairwise decisions between songs in this shared subset, ensuring that both modalities are compared over the same candidates. At the model level, we compute Spearman rank correlation between each ALM and VLM ranking. At the consensus level, we construct separate audio and visual rankings by averaging candidate ranks across ALMs and VLMs, respectively, and re-ranking the candidates. We then compute Spearman rank correlation between the two consensus rankings.

\section{Experiments}
\label{sec:experiments}

\begin{table}[t]
\centering
\small
\begin{tabular}{l cc}
\toprule
\textbf{Group} & \textbf{Kendall's $W$} & \textbf{$p$-value} \\
\midrule
RED VELVET     & 0.334 & $<0.001$ \\
TWICE          & 0.240 & 0.043 \\
BLACKPINK      & 0.319 & $<0.001$ \\
I-DLE          & 0.356 & $<0.001$ \\
ITZY           & 0.272 & 0.010 \\
AESPA          & 0.263 & 0.015 \\
IVE            & 0.419 & $<0.001$ \\
NMIXX          & 0.297 & 0.003 \\
LE SSERAFIM    & 0.314 & $<0.001$ \\
NEWJEANS       & 0.375 & $<0.001$ \\
BABYMONSTER    & 0.226 & 0.104 \\
ILLIT          & \textbf{0.427} & $<0.001$ \\
MEOVV          & 0.191 & 0.321 \\
HEARTS2HEARTS  & 0.137 & 0.675 \\
\bottomrule
\end{tabular}
\caption{
Group-level aesthetic consensus among ALMs. Kendall's coefficient of concordance $W$ is computed from the rankings of all six model configurations for each artist group in the audio-based evaluation.
}
\vspace{-3mm}
\label{tab:audio_kendall}
\end{table}

\subsection{Experimental Setup}
\label{sec:experimental_setup}

\paragraph{Dataset Construction.}
We construct our evaluation dataset from 14 K-pop girl groups spanning different debut periods and artist identities. For each group, we first collect songs with official music videos, with at most 30 videos retained per group. The corresponding audio tracks of these songs are included in the audio evaluation set, ensuring that the audio and video settings share a substantial set of common content.

Because many groups have fewer than 30 eligible official music videos, we further expand the audio collection with additional popular songs from the same artist, up to a maximum of 30 audio tracks per group. For groups whose available discography contains fewer than 30 eligible songs, we retain all available tracks. As a result, the video set forms a subset of the corresponding audio collection for each artist, while the audio setting generally contains a broader selection of songs.

The resulting dataset contains 373 audio tracks and 278 music videos. Detailed statistics for each artist group are reported in Table~\ref{tab:dataset_statistics}, with the complete candidate lists provided in Appendix~\ref{app:song_list}.

% \begin{table}[t]
% \centering
% \small
% \begin{tabular}{l cc}
% \toprule
% \textbf{Group} & \textbf{Kendall's $W$} & \textbf{$p$-value} \\
% \midrule
% RED VELVET     & 0.334 & 0.0002 \\
% TWICE          & 0.240 & 0.0425 \\
% BLACKPINK      & 0.319 & 0.0008 \\
% I-DLE          & 0.356 & $<0.001$ \\
% ITZY           & 0.272 & 0.0101 \\
% AESPA          & 0.263 & 0.0148 \\
% IVE            & 0.419 & $<0.001$ \\
% NMIXX          & 0.297 & 0.0027 \\
% LE SSERAFIM    & 0.314 & 0.0008 \\
% NEWJEANS       & 0.375 & $<0.001$ \\
% BABYMONSTER    & 0.226 & 0.1040 \\
% ILLIT          & \textbf{0.427} & $<0.001$ \\
% MEOVV          & 0.191 & 0.3208 \\
% HEARTS2HEARTS  & 0.137 & 0.6752 \\
% \bottomrule
% \end{tabular}
% \caption{
% Group-level aesthetic consensus among ALMs. Kendall's coefficient of concordance $W$ is computed from the rankings of all six model configurations for each artist group in the audio-based evaluation.
% }
% \vspace{-3mm}
% \label{tab:audio_kendall}
% \end{table}

\paragraph{Audio Models.}
For audio-based evaluation, we consider six audio-language models (ALMs) from two audio-capable foundation model families: MOSS-4B-Instruct~\cite{yang2026moss}, MOSS-4B-Thinking~\cite{yang2026moss}, MOSS-8B-Instruct~\cite{yang2026moss}, MOSS-8B-Thinking~\cite{yang2026moss}, Qwen3-Omni-30B-A3B-Instruct~\cite{xu2025qwen3}, and Qwen3-Omni-30B-A3B-Thinking~\cite{xu2025qwen3}. We select these models based on both their strong audio understanding capabilities and their support for processing multiple audio inputs within a single context, which enables direct pairwise comparison between two songs.

% \paragraph{Video Models.}
% For music-video evaluation, we use the Qwen3.5~\cite{qwen3.5} family at three model scales: Qwen3.5-2B, Qwen3.5-4B, Qwen3.5-9B, and Qwen3.5-27B, InternVL3.5~\cite{wang2025internvl3} family: 2B, 4B, 8B, 14B.  Each model is evaluated under both non-thinking and thinking modes, resulting in six model configurations in total. This controlled design allows us to study the effects of both model scale and reasoning mode while keeping the underlying model family fixed.

\begin{figure}[t]
    \centering
    \includegraphics[width=\linewidth]{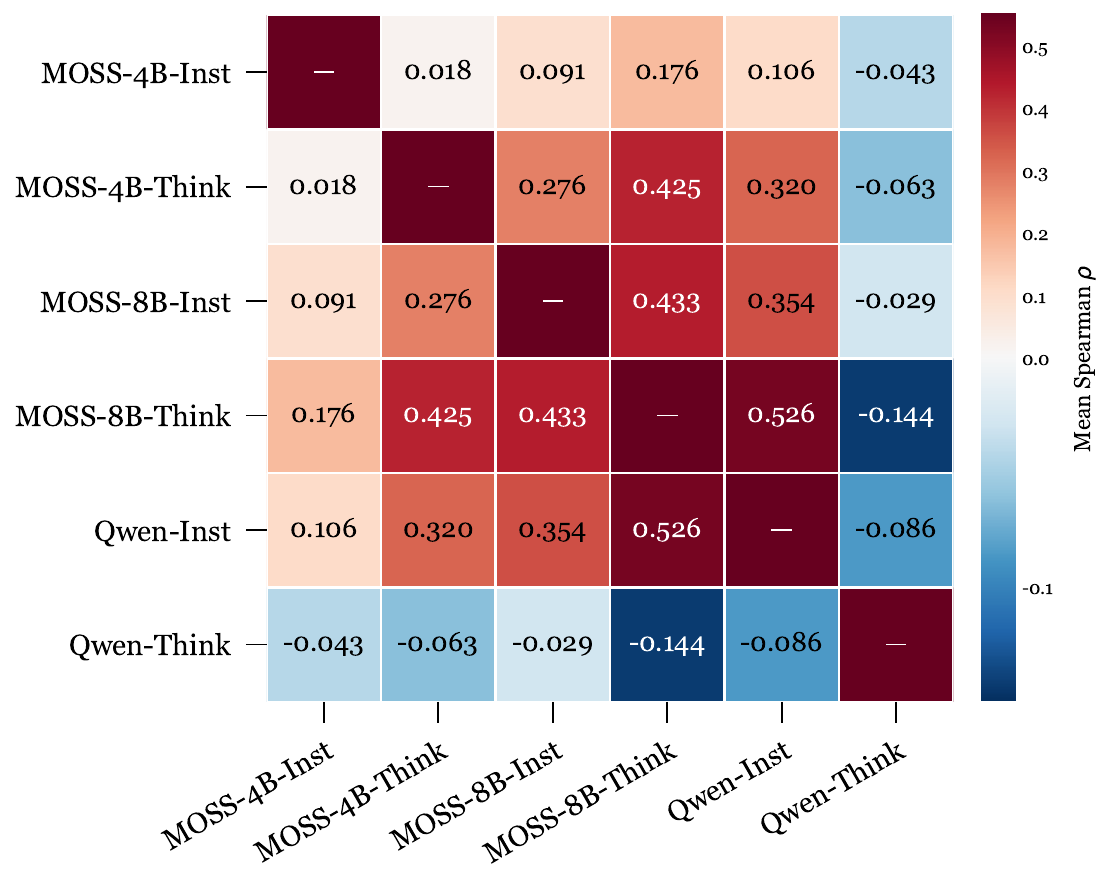}
    \caption{
    Pairwise ranking agreement between ALMs. Each cell reports the mean Spearman rank correlation across 14 artist groups. Inst and Think denote Instruct and Thinking configurations, respectively; Qwen refers to Qwen3-Omni-30B-A3B.
    }
    \label{fig:audio_spearman}
    \vspace{-3mm}
\end{figure}

\paragraph{Video Models.}
For music-video evaluation, we consider 28 model configurations from five vision-language model (VLM) families. Specifically, we evaluate Qwen3.5~\cite{qwen3.5} at four model scales (2B, 4B, 9B, and 27B), InternVL3.5~\cite{wang2025internvl3} at six scales (1B, 2B, 4B, 8B, 14B, and 38B), GLM-4.6V-Flash~\cite{hong2025glm}, and MiniCPM-V-4.5~\cite{yu2026minicpm}. For Qwen3.5, InternVL3.5, GLM-4.6V-Flash, and MiniCPM-V-4.5, we evaluate both non-thinking and thinking modes whenever supported. We additionally include Qwen3-VL-Instruct~\cite{bai2025qwen3} at four model scales (2B, 4B, 8B, and 32B). This diverse model set allows us to examine aesthetic agreement across model families, model scales, and inference modes.

\paragraph{Evaluation Protocol.}
For both modalities, comparisons are performed only among candidates belonging to the same artist group. ALMs receive only the audio content, while VLMs receive only the visual content of the corresponding music videos. Song titles, chart performance, popularity statistics, and other external metadata are excluded from the model input.

All candidate pairs are evaluated under both presentation orders following the pairwise ranking procedure described in Section~\ref{sec:pairwise_ranking}. To reduce the impact of randomness, each pairwise comparison is independently repeated five times under each presentation order. The resulting preferences are aggregated to construct within-group rankings, which are analyzed using the aesthetic agreement metrics defined in Section~\ref{sec:agreement}. For cross-modal analysis, we restrict the comparison to songs shared by both modalities and recompute the audio rankings using pairwise decisions within this shared subset, following Section~\ref{sec:evaluation_settings}. We use the same evaluation prompt and decoding configuration across candidates within each model setting.

\begin{table}[t]
\centering
\small
\begin{tabular}{l cc}
\toprule
\textbf{Group} & \textbf{Kendall's $W$} & \textbf{$p$-value} \\
\midrule
RED VELVET    & 0.656 & $<0.001$ \\
TWICE          & 0.649 & $<0.001$ \\
BLACKPINK      & \textbf{0.769} & $<0.001$ \\
I-DLE          & 0.681 & $<0.001$ \\
ITZY           & 0.516 & $<0.001$ \\
AESPA          & 0.584 & $<0.001$ \\
IVE            & 0.692 & $<0.001$ \\
NMIXX          & 0.555 & $<0.001$ \\
LE SSERAFIM    & 0.503 & $<0.001$ \\
NEWJEANS       & 0.536 & $<0.001$ \\
BABYMONSTER    & 0.506 & $<0.001$ \\
ILLIT          & 0.470 & $<0.001$ \\
MEOVV          & 0.532 & $<0.001$ \\
HEARTS2HEARTS  & 0.557 & $<0.001$ \\
\bottomrule
\end{tabular}
\caption{
Group-level aesthetic consensus among VLMs. Kendall's coefficient of concordance $W$ is computed from the rankings of all 28 model configurations for each artist group in the music-video setting.
}
\vspace{-3mm}
\label{tab:video_kendall}
\end{table}

\subsection{Experimental Results}

\paragraph{Audio-based Aesthetic Agreement.}
We first examine whether ALMs exhibit similar aesthetic preferences when evaluating songs based solely on their audio content. Figure~\ref{fig:audio_spearman} presents the pairwise Spearman correlations averaged across 14 artist groups. Overall, agreement is limited, with the strongest model pair reaching a correlation of only $0.526$, while several pairs exhibit near-zero or negative correlations. Notably, Qwen3-Omni-Thinking shows consistently weak agreement with the other configurations, illustrating substantial variation in musical preferences across models.

We further assess group-level consensus using Kendall's $W$ (shown in Table~\ref{tab:audio_kendall}). Concordance ranges from $0.137$ to $0.427$, with 11 of the 14 groups showing statistically significant agreement ($p<0.05$). Although the overall consensus remains relatively modest, these results indicate that shared ranking tendencies can still emerge within individual artist groups despite considerable differences between models.

\vspace{-1mm}
\paragraph{Visual Aesthetic Agreement.}

\begin{figure*}[t]
    \centering
    \includegraphics[width=\textwidth]{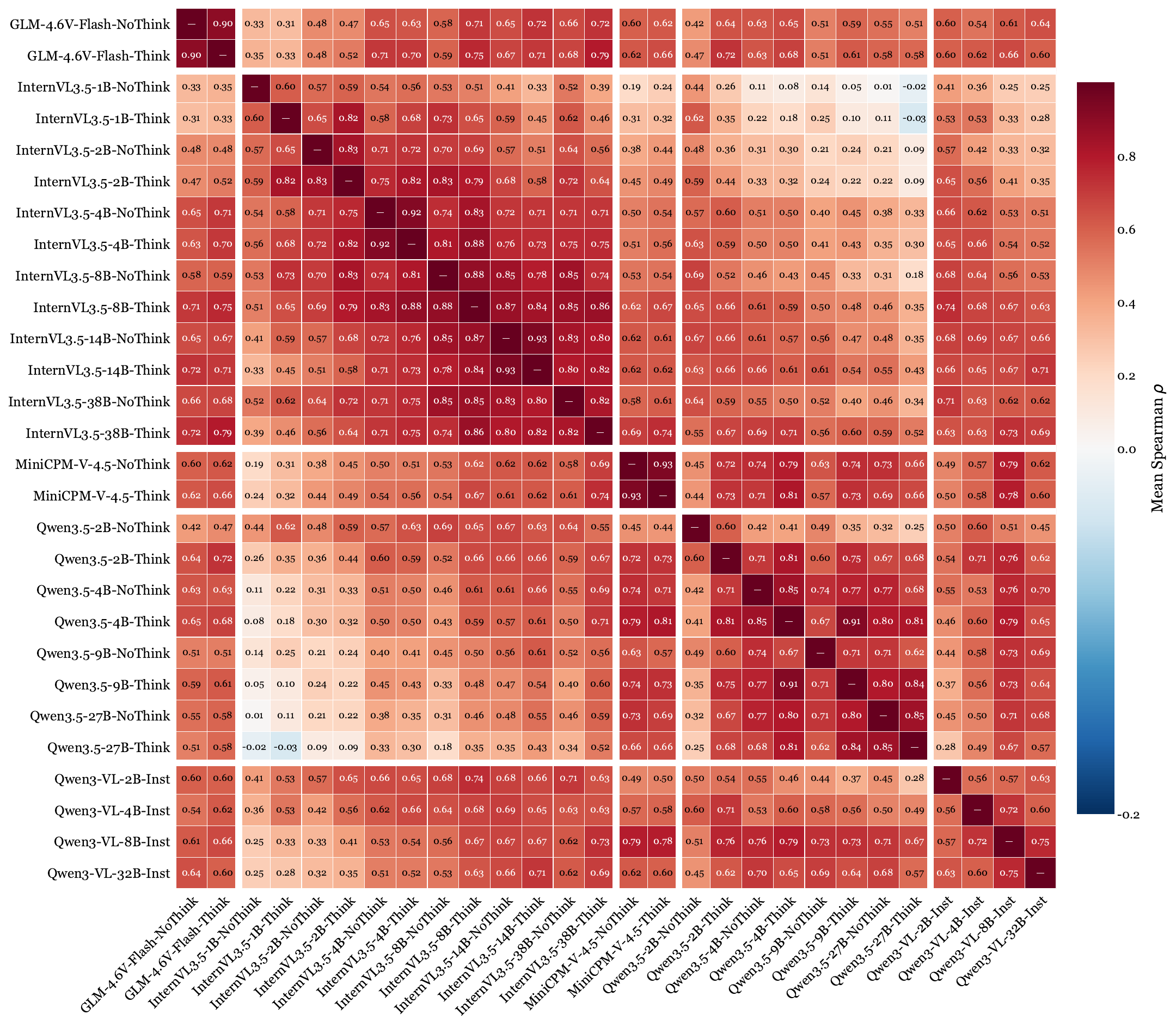}
    \caption{
    Pairwise ranking agreement between VLMs. Each cell reports the mean Spearman rank correlation across 14 artist groups in the music-video setting. Models are grouped by family, showing strong within-family agreement and substantial cross-family correlations.
    }
    \label{fig:video_spearman}
    \vspace{-3mm}
\end{figure*}

We next examine aesthetic agreement among VLMs in the music-video setting. As shown in Figure~\ref{fig:video_spearman}, pairwise Spearman correlations are substantially higher than those observed in audio-based evaluation. Strong positive correlations appear not only within the same model family but also across independently developed families, suggesting broadly shared visual preferences. Meanwhile, the visible block structure indicates that model family and inference configuration still influence the resulting rankings.

\begin{table}[t]
\centering
\small
\begin{tabular}{l cc}
\toprule
\textbf{Group} & \textbf{Spearman $\rho$} & \textbf{$p$-value} \\
\midrule
Red Velvet     & 0.023 & 0.905 \\
TWICE          & \textbf{0.475} & \textbf{0.007} \\
BLACKPINK      & 0.178 & 0.344 \\
I-DLE          & 0.377 & 0.064 \\
ITZY           & 0.356 & 0.074 \\
AESPA          & 0.006 & 0.977 \\
IVE            & 0.279 & 0.193 \\
NMIXX          & -0.219 & 0.413 \\
LE SSERAFIM    & -0.010 & 0.972 \\
NewJeans       & 0.177 & 0.512 \\
BABYMONSTER    & 0.399 & 0.128 \\
ILLIT          & -0.084 & 0.793 \\
MEOVV          & 0.107 & 0.840 \\
HEARTS2HEARTS  & 0.167 & 0.680 \\
\bottomrule
\end{tabular}
\caption{
Cross-modal consensus alignment between audio and visual evaluations. Spearman rank correlation is computed between the audio and visual consensus rankings for each artist group, using only songs shared by both modalities.}
\vspace{-3mm}
\label{tab:crossmodal_consensus}
\end{table}

This stronger agreement is also reflected in the group-level analysis (shown in Table~\ref{tab:video_kendall}), where Kendall's $W$ ranges from $0.470$ to $0.769$. All 14 artist groups exhibit statistically significant concordance ($p<0.001$), with the highest agreement observed for BLACKPINK ($W=0.769$) and the lowest for ILLIT ($W=0.470$). Together, these findings indicate that visual aesthetic preferences are more consistently shared among the evaluated models than audio-based preferences, although model-specific differences remain.

\begin{figure*}[t]
    \centering
    \includegraphics[width=\textwidth]{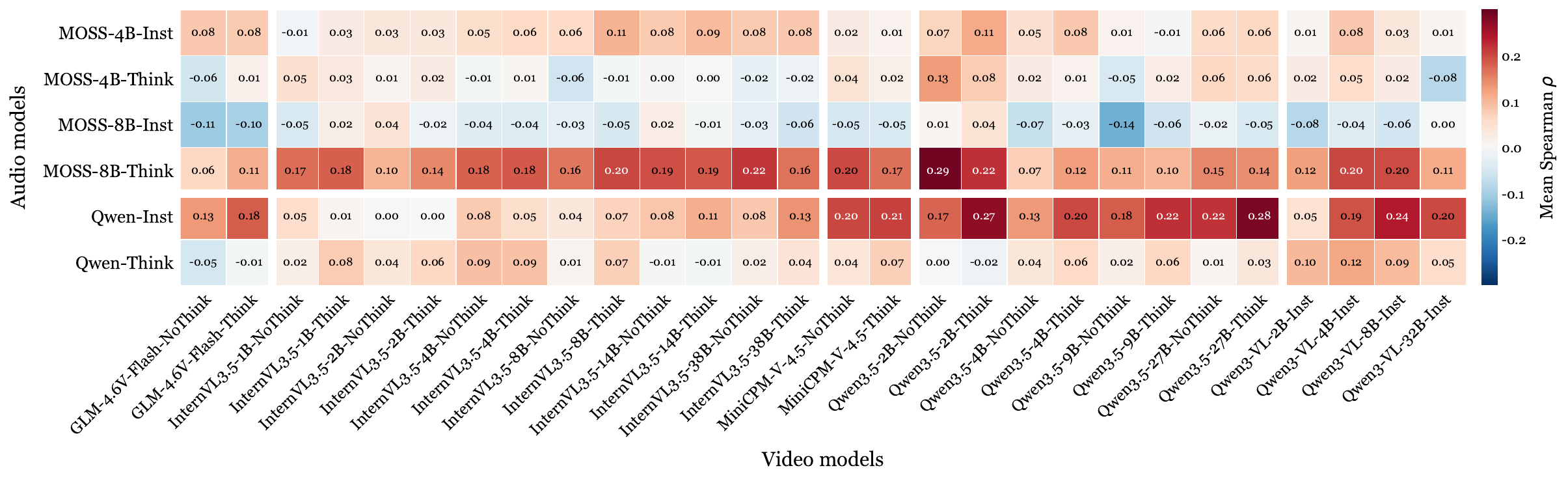}
    \caption{
    Cross-modal ranking agreement between ALMs and VLMs. Each cell reports the mean Spearman rank correlation across 14 artist groups, computed using only songs shared by both evaluation sets.
    }
\vspace{-4mm}
\label{fig:crossmodal_spearman}
\end{figure*}

\vspace{-1mm}
\paragraph{Cross-modal Preference Alignment.}
We next examine whether aesthetic preferences derived from audio remain aligned with those obtained from the corresponding music videos. For each artist group, we restrict the analysis to songs shared by both evaluation sets and compute the Spearman correlation between each audio-model ranking and each video-model ranking. Figure~\ref{fig:crossmodal_spearman} reports the resulting correlations averaged across 14 artist groups.

Overall, cross-modal agreement is weak, with a mean Spearman correlation of only $0.060$ across all audio--video model pairs. Most correlations are close to zero, and even the strongest model pair reaches only $0.288$. This indicates that audio-based preferences are generally not preserved in visual evaluation.

% We further compare the audio and visual consensus rankings for each artist group (shown in Table~\ref{tab:crossmodal_consensus}). The correlations range from $-0.219$ to $0.475$, with only TWICE showing statistically significant alignment ($\rho=0.475$, $p=0.007$). I-DLE ($\rho=0.377$) and ITZY ($\rho=0.356$) exhibit relatively higher positive correlations, although neither reaches statistical significance ($p<0.05$). Several other groups show near-zero or negative correlations.

We further compare the audio and visual consensus rankings for each artist group (shown in Table~\ref{tab:crossmodal_consensus}). The correlations range from $-0.219$ to $0.475$, with only TWICE showing statistically significant alignment ($\rho=0.475$, $p=0.007$). BABYMONSTER ($\rho=0.399$), I-DLE ($\rho=0.377$), and ITZY ($\rho=0.356$) also exhibit relatively higher positive correlations, although none reaches statistical significance ($p<0.05$). Several other groups show near-zero or negative correlations.

Together, these results reveal a clear contrast: although visual evaluation yields substantially stronger inter-model consensus, this agreement does not necessarily extend to audio-based preferences. The two evaluation settings therefore exhibit largely distinct aesthetic preference patterns.

\vspace{-1.5mm}
\paragraph{Alignment with Real-world Popularity.}
\label{sec:market_alignment}

Finally, we examine how model-derived visual preferences relate to real-world popularity. We compare VLM consensus rankings with historical YouTube view counts for official MVs from AESPA, IVE, and TWICE. Using Internet Archive snapshots taken approximately seven days after release ($\pm 2$ days), we obtain historical view counts for 64 of the 76 MVs and compute Spearman rank correlations within each artist group. Further details are provided in Appendix~\ref{app:market_alignment}.

\begin{table}[t]
\centering
\small
\setlength{\tabcolsep}{5pt}
\renewcommand{\arraystretch}{1.1}
\begin{tabular}{lccc}
\toprule
\textbf{Artist} & \textbf{\# MVs} & \textbf{Spearman $\rho$} & \textbf{$p$-value} \\
\midrule
AESPA & 19/23 & \textbf{0.661} & \textbf{0.002} \\
IVE   & 19/23 & 0.363 & 0.127 \\
TWICE & 26/30 & 0.406 & 0.040 \\
\bottomrule
\end{tabular}
\caption{
Association between VLM consensus rankings and early YouTube viewership. Historical view counts are collected approximately seven days after release ($\pm 2$ days). We report within-group Spearman correlations and two-sided $p$-values.
}
\vspace{-5mm}
\label{tab:market_alignment}
\end{table}

As shown in Table~\ref{tab:market_alignment}, all three groups exhibit positive correlations, with the strongest association observed for AESPA ($\rho=0.661$), followed by TWICE ($\rho=0.406$) and IVE ($\rho=0.363$). Only AESPA remains statistically significant after correcting for multiple comparisons. Similar positive trends are observed under alternative observation windows. These findings suggest a possible association between machine visual preferences and early YouTube viewership. 
% \section{Discussion}
% This study represents an initial step toward AI-assisted creative decision making in the K-pop industry. By examining how foundation models evaluate K-pop songs and music videos, we begin to understand the role of machine aesthetic preferences in future creative decision-making systems. Our findings suggest that creative preferences cannot be fully captured by a single measure of quality, motivating a broader perspective that considers factors beyond aesthetic evaluation alone.

% Building on this initial understanding, our long-term goal is to progressively incorporate artist identity, audience preferences, historical performance, and market dynamics into a unified framework. Ultimately, we envision a K-pop world model that captures how artists, audiences, and the broader market interact and evolve, enabling AI systems to explore potential market responses to different creative strategies. Such a system could support song selection, comeback planning, artist positioning, and long-term strategic development.

\vspace{-1mm}
\section{Discussion}
\vspace{-1mm}

This study represents an initial step toward AI-assisted creative decision making in the K-pop industry. By examining how foundation models evaluate K-pop songs and music videos, we begin to understand the role of machine aesthetic preferences in future creative decision-making systems. Our findings reveal differences in aesthetic agreement across audio and visual evaluation, as well as limited alignment between the two modalities. These observations suggest that machine taste cannot be fully characterized by a single aesthetic perspective. Moreover, the varying relationships between visual preferences and real-world popularity highlight the importance of considering factors beyond model-derived aesthetic evaluations when studying creative content in real-world settings.

Building on this initial understanding, our long-term goal is to progressively integrate machine aesthetic preferences with artist identity, audience feedback, historical performance, and market dynamics into a unified framework. Ultimately, we envision a K-pop world model that captures how creative content, artists, audiences, and the broader market interact and evolve, enabling AI systems to explore potential market responses to different creative strategies. Rather than treating aesthetic evaluation or historical popularity as isolated indicators, such a framework could provide a broader understanding of how creative choices interact with artist characteristics and changing market conditions. In the long term, this could support tasks such as song selection, comeback planning, artist positioning, and strategic development, while complementing the expertise of creative professionals.

% \section{Conclusion}

% In this work, we investigate machine aesthetic preferences through pairwise evaluations of K-pop songs and music videos. Our results reveal substantially stronger inter-model agreement in visual evaluation than in audio evaluation, while preferences across the two modalities remain weakly aligned. We further observe varying degrees of association between visual consensus rankings and early real-world audience reception. Together, these findings provide an empirical understanding of how foundation models evaluate subjective creative content.

\vspace{-1mm}
\section{Conclusion}
\vspace{-1mm}

In this work, we investigate machine aesthetic preferences through pairwise evaluations of K-pop songs and music videos. Visual evaluation shows stronger inter-model agreement and group-level consensus than audio evaluation, while cross-modal preferences remain weakly aligned. We also find positive but varying correlations between visual consensus rankings and historical YouTube viewership. These findings offer initial insights into machine taste and future AI-assisted creative decision making.

\section*{Limitations}

Our study focuses on 14 K-pop girl groups and a selected set of foundation models, so the findings may not fully generalize to other creative domains. The audio and video evaluations also involve different model populations, and the video setting considers visual content only. In addition, YouTube viewership provides a proxy for real-world popularity rather than direct human aesthetic judgments. Future work could extend the evaluation to broader datasets, additional model families, and human preference studies.

\section*{Ethical Considerations}
Our study examines aesthetic preferences expressed by foundation models rather than establishing objective standards of creative quality. Model-derived rankings may reflect biases in training data and should not be treated as substitutes for human aesthetic judgment. In future AI-assisted creative decision-making systems, excessive reliance on such rankings could reinforce existing popularity patterns and potentially reduce creative diversity. We therefore emphasize that machine aesthetic preferences should complement, rather than replace, the perspectives of artists, audiences, and creative professionals.

\bibliography{custom}

\appendix

\newpage

% \newpage
\section*{Appendix}

\section{Data Usage and Copyright}
Our evaluation uses publicly released K-pop songs and official music videos solely for academic research. All original audio and video content remains the property of its respective copyright holders.

\section{Full Song List}
\label{app:song_list}

Tables~\ref{tab:full_audio_list} and~\ref{tab:full_mv_list} provide the complete lists of songs and music videos used in our audio-based and video-based evaluations, respectively, covering all 14 K-pop artist groups.

\begin{table*}[!t]
\centering

\resizebox{\linewidth}{!}{
\begin{tabular}{lcccccc}
\toprule
\textbf{Model}
& \textbf{MOSS-4B-Inst}
& \textbf{MOSS-4B-Think}
& \textbf{MOSS-8B-Inst}
& \textbf{MOSS-8B-Think}
& \textbf{Qwen-Inst}
& \textbf{Qwen-Think} \\
\midrule
MOSS-4B-Inst
& --      & 0.018  & 0.091  & 0.176  & 0.106  & -0.043 \\

MOSS-4B-Think
&         & --     & 0.276  & 0.425  & 0.320  & -0.063 \\

MOSS-8B-Inst
&         &        & --     & 0.433  & 0.354  & -0.029 \\

MOSS-8B-Think
&         &        &        & --     & \textbf{0.526}  & -0.144 \\

Qwen-Inst
&         &        &        &        & --     & -0.086 \\

Qwen-Think
&         &        &        &        &        & -- \\
\bottomrule
\end{tabular}
}
\caption{
Pairwise ranking agreement between ALMs. Each cell reports the mean Spearman rank correlation across 14 artist groups. Higher values indicate stronger agreement between model rankings.
}
\label{tab:audio_model_spearman}
\end{table*}

\section{Detailed Experimental Results}
\label{app:detailed_results}

We provide the complete pairwise Spearman correlation results underlying the agreement heatmaps in the main paper. Table~\ref{tab:audio_model_spearman} reports the audio-based results, while Table~\ref{tab:video_spearman_full} presents the music-video-based results, with model abbreviations listed in Table~\ref{tab:video_model_abbrev}.

\begin{table}[t]
\centering
\small
\resizebox{0.98\linewidth}{!}{
\begin{tabular}{ll}
\toprule
\textbf{Abbreviation} & \textbf{Model} \\
\midrule
G-N / G-T & GLM-4.6V-Flash, non-thinking / thinking \\
I1-N / I1-T & InternVL3.5-1B, non-thinking / thinking \\
I2-N / I2-T & InternVL3.5-2B, non-thinking / thinking \\
I4-N / I4-T & InternVL3.5-4B, non-thinking / thinking \\
I8-N / I8-T & InternVL3.5-8B, non-thinking / thinking \\
I14-N / I14-T & InternVL3.5-14B, non-thinking / thinking \\
I38-N / I38-T & InternVL3.5-38B, non-thinking / thinking \\
M-N / M-T & MiniCPM-V-4.5, non-thinking / thinking \\
Q2-N / Q2-T & Qwen3.5-2B, non-thinking / thinking \\
Q4-N / Q4-T & Qwen3.5-4B, non-thinking / thinking \\
Q9-N / Q9-T & Qwen3.5-9B, non-thinking / thinking \\
Q27-N / Q27-T & Qwen3.5-27B, non-thinking / thinking \\
V2 / V4 / V8 / V32 & Qwen3-VL-2B/4B/8B/32B-Instruct \\
\bottomrule
\end{tabular}}
\caption{Abbreviations used for VLMs in Table~\ref{tab:video_spearman_full}.}
\label{tab:video_model_abbrev}
\end{table}

\section{Prompts}
\label{app:prompts}
We provide the complete system and question prompts used for audio-based song ranking and music-video-based evaluation. The placeholder \texttt{\{group\}} is replaced with the corresponding artist group name during evaluation.

\subsection{Audio Evaluation}

\begin{tcolorbox}[
    colback=gray!3,
    colframe=gray!45,
    boxrule=0.5pt,
    arc=2pt,
    title=\textbf{System Prompt},
    fonttitle=\small,
    breakable
]
\small

You are a highly experienced K-pop music critic and a devoted, long-time fan of \texttt{\{group\}}. You are comparing two songs by \texttt{\{group\}} and must judge which one is the BETTER SONG based on musical quality and craftsmanship -- not simply which one gives you the strongest nostalgia or hype in the moment.

\medskip
\textbf{Important instructions:}

\begin{itemize}[leftmargin=*, nosep]
    \item Track labels A and B are arbitrary and carry no meaning.
    \item Playback order must NOT influence your judgment.
    \item Evaluate each track independently and carefully before comparing them. Pay close attention to concrete musical quality dimensions you can actually judge from listening: melody and hook strength, vocal performance and harmonies, production and mixing quality, arrangement and song structure, and overall cohesion. Do not judge originality -- you cannot reliably tell from a single listen whether something is derivative of other songs, so do not factor that in.
    \item Do not favor the first or second track simply because of presentation order.
    \item Base your judgment strictly on the audio quality and craftsmanship you actually hear in each track, not on which one you recognize, remember more vividly, or assume is more popular.
\end{itemize}

\medskip
Do not identify or guess the song titles. Base your judgment only on the audio.

\end{tcolorbox}

\begin{tcolorbox}[
    colback=gray!3,
    colframe=gray!45,
    boxrule=0.5pt,
    arc=2pt,
    title=\textbf{Question Prompt},
    fonttitle=\small,
    breakable
]
\small

Having carefully and independently evaluated the musical quality of Track A and Track B -- melody, vocals, production, and arrangement -- which track is the better song overall?

\medskip
Answer with EXACTLY one line and nothing else:

\medskip
\texttt{PREFERENCE: A}

or

\texttt{PREFERENCE: B}

\end{tcolorbox}

\subsection{Music-Video Evaluation}

\begin{tcolorbox}[
    colback=gray!3,
    colframe=gray!45,
    boxrule=0.5pt,
    arc=2pt,
    title=\textbf{System Prompt},
    fonttitle=\small,
    breakable
]
\small

You are a highly experienced K-pop music video critic and a devoted, long-time fan of \texttt{\{group\}}. You are comparing two official music videos by \texttt{\{group\}} and must judge which one is the BETTER MUSIC VIDEO as a visual production -- not simply which one gives you the strongest nostalgia or hype in the moment, and not based on the song itself.

\medskip
\textbf{Important instructions:}

\begin{itemize}[leftmargin=*, nosep]
    \item Video labels A and B are arbitrary and carry no meaning.
    \item Playback order must NOT influence your judgment.
    \item Evaluate each video independently and carefully before comparing them. Pay close attention to concrete visual quality dimensions you can actually judge from watching: choreography and dance performance, cinematography and camera work, visual concept and styling (costumes, sets, color grading), editing and pacing, and overall visual cohesion and production value. Do not judge originality -- you cannot reliably tell from a single viewing whether something is derivative of other MVs, so do not factor that in.
    \item Do not favor the first or second video simply because of presentation order.
    \item Base your judgment strictly on the visual quality and craftsmanship you actually see in each video, not on which one you recognize, remember more vividly, or assume is more popular.
\end{itemize}

\medskip
Do not identify or guess the song titles. Base your judgment only on what you see.

\end{tcolorbox}

\begin{tcolorbox}[
    colback=gray!3,
    colframe=gray!45,
    boxrule=0.5pt,
    arc=2pt,
    title=\textbf{Question Prompt},
    fonttitle=\small,
    breakable
]
\small

Having carefully and independently evaluated the visual quality of Video A and Video B -- choreography, cinematography, styling/concept, and editing -- which video is the better music video overall?

\medskip
Answer with EXACTLY one line and nothing else:

\medskip
\texttt{PREFERENCE: A}

or

\texttt{PREFERENCE: B}

\end{tcolorbox}

\section{Case Studies}
\label{app:case_studies}

To complement the aggregate results, we present three song-level case studies corresponding to the main evaluation settings. We examine inter-model agreement in AESPA's audio evaluation, group-level consensus in IVE's music-video evaluation, and cross-modal preference alignment in TWICE. All reported rankings are derived from the original pairwise evaluation results, where a lower rank indicates a stronger preference.

\subsection{Inter-model Agreement: AESPA}

\begin{table*}[t]
\centering
\small
\setlength{\tabcolsep}{5pt}
\renewcommand{\arraystretch}{1.12}
\begin{tabular}{lrrrrrrr}
\toprule
\textbf{Song}
& \textbf{M4-I}
& \textbf{M4-T}
& \textbf{M8-I}
& \textbf{M8-T}
& \textbf{Q-I}
& \textbf{Q-T}
& \textbf{Consensus} \\
\midrule
Savage              & 10 & 2  & 2  & 1  & 6  & 16 & 1  \\
Welcome To MY World & 7  & 3  & 1  & 5  & 5  & 24 & 2  \\
Drama               & 5  & 14 & 8  & 2  & 3  & 23 & 3  \\
Count On Me         & 15 & 10 & 16 & 6  & 15 & 6  & 6  \\
Spicy               & 9  & 25 & 10 & 2  & 2  & 24 & 8  \\
Next Level          & 27 & 4  & 6  & 9  & 12 & 27 & 13 \\
Better Things       & 30 & 9  & 2  & 18 & 24 & 7  & 16 \\
Hold On Tight       & 2  & 30 & 15 & 30 & 21 & 13 & 23 \\
Black Mamba         & 21 & 26 & 23 & 21 & 8  & 29 & 28 \\
Can't Help Myself   & 28 & 19 & 26 & 23 & 22 & 30 & 30 \\
\bottomrule
\end{tabular}

\caption{
Song-level audio rankings of all 30 AESPA songs across six ALM configurations. M4, M8, and Q denote MOSS-Audio-4B, MOSS-Audio-8B, and Qwen3-Omni-30B-A3B, respectively; I and T indicate Instruct and Thinking. Consensus rankings are obtained by averaging the six model rankings and re-ranking the songs. Lower ranks indicate stronger preferences.
}

\label{tab:aespa_intermodel_case}
\end{table*}

Table~\ref{tab:aespa_intermodel_case} illustrates substantial variation in the audio preferences of individual models. Savage receives favorable rankings from most configurations, particularly MOSS-Audio-8B-Thinking (rank 1), although Qwen3-Omni-Thinking places it at rank 16. The disagreement is more pronounced for several other songs. For example, Hold On Tight ranks second under MOSS-Audio-4B-Instruct but last under both MOSS-Audio-4B-Thinking and MOSS-Audio-8B-Thinking. Similarly, Next Level is ranked fourth by MOSS-Audio-4B-Thinking but 27th by MOSS-Audio-4B-Instruct and Qwen3-Omni-Thinking. In contrast,
Can't Help Myself receives consistently low rankings across all six configurations. These examples illustrate how individual model preferences can differ substantially, even when evaluating the same songs under an identical protocol.

\subsection{Group-level Consensus: IVE}

\begin{table}[t]
\centering
\small
\setlength{\tabcolsep}{5pt}
\renewcommand{\arraystretch}{1.12}
\begin{tabular}{lrrr}
\toprule
\textbf{Song}
& \textbf{Mean Rank}
& \textbf{Rank SD}
& \textbf{Consensus} \\
\midrule
HEYA           & 2.54  & 1.82 & 1  \\
I AM           & 4.50  & 2.79 & 2  \\
ELEVEN         & 4.55  & 3.62 & 3  \\
BLACKHOLE      & 5.39  & 4.49 & 4  \\
LOVE DIVE      & 5.71  & 2.82 & 5  \\
BANG BANG      & 7.46  & 4.02 & 6  \\
After LIKE     & 8.27  & 3.35 & 9  \\
Accendio       & 13.77 & 4.01 & 14 \\
Off The Record & 19.00 & 3.82 & 21 \\
Either Way     & 20.46 & 1.85 & 23 \\
\bottomrule
\end{tabular}

\caption{
Selected IVE songs illustrating visual consensus among 28 VLM configurations over 23 music videos. Mean Rank and Rank SD denote the mean and sample standard deviation of model rankings, respectively. Consensus is obtained by re-ranking the mean ranks. Lower ranks indicate stronger preferences, while smaller SD values indicate greater agreement.
}

\label{tab:ive_consensus_case}
\end{table}

Table~\ref{tab:ive_consensus_case} presents representative consensus rankings from IVE's visual evaluation. HEYA receives the highest consensus ranking, with a mean rank of 2.54 and a relatively small standard deviation of 1.82, indicating that most evaluated configurations place it near the top. I AM and LOVE DIVE also receive favorable rankings with comparatively limited variation. In contrast, BLACKHOLE ranks fourth overall but exhibits greater dispersion (SD = 4.49), suggesting less agreement on its exact position despite its favorable average ranking. At the other end, Either Way receives the lowest consensus ranking with a similarly small SD of 1.85. These examples illustrate that group-level consensus can capture both shared high and low preferences, while individual songs may still exhibit different degrees of inter-model variation.

\subsection{Cross-modal Preference Alignment: TWICE}

\begin{table}[t]
\centering
\small
\setlength{\tabcolsep}{5pt}
\renewcommand{\arraystretch}{1.12}
\begin{tabular}{lrr}
\toprule
\textbf{Song}
& \textbf{Audio Rank}
& \textbf{Visual Rank} \\
\midrule
Fancy        & 2  & 2  \\
Feel Special & 5  & 5  \\
Celebrate    & 3  & 6  \\
Heart Shaker & 17 & 17 \\
TT           & 25 & 24 \\
Cheer Up     & 30 & 29 \\
More \& More & 10 & 1  \\
Scientist    & 23 & 9  \\
YES or YES   & 1  & 20 \\
Doughnut     & 6  & 26 \\
\bottomrule
\end{tabular}

\caption{
Selected TWICE songs illustrating cross-modal preference alignment. Audio and visual consensus rankings are obtained from six ALM and 28 VLM configurations, respectively, using the same 30 songs with official music videos. Lower ranks indicate stronger preferences.
}

\label{tab:twice_crossmodal_case}
\end{table}

TWICE exhibits relatively stronger cross-modal consensus alignment ($\rho=0.475$), as illustrated in Table~\ref{tab:twice_crossmodal_case}. Fancy and Feel Special receive identical rankings across audio and visual evaluations, while Celebrate also ranks highly in both. Similar agreement appears among lower-ranked songs, including TT and Cheer Up. Nevertheless, notable differences remain. More \& More ranks first visually but tenth in audio, while Scientist receives a considerably higher visual ranking (9 versus 23). Conversely, YES or YES ranks first in audio but twentieth visually, and Doughnut ranks sixth in audio but 26th visually. These examples demonstrate that stronger aggregate cross-modal agreement does not necessarily imply consistent preferences for every individual song.

\section{Real-world Popularity Analysis}
\label{app:market_alignment}

\paragraph{Data Collection.}
To examine the relationship between model-derived aesthetic preferences and real-world popularity, we collect historical YouTube view counts for official music videos from AESPA, IVE, and TWICE. These three groups contain 23, 23, and 30 music videos in our evaluation dataset, respectively, resulting in 76 videos in total.

We first match each song to its corresponding official YouTube video. Candidate videos are identified through Kworb and web search, and subsequently verified using their titles and uploading channels. We retain official music videos while excluding alternative content such as dance practices, performance videos, lyric videos, teasers, and remixes. All 76 music videos are successfully matched to their official YouTube video IDs.

Since the YouTube Data API provides current cumulative view counts but does not offer publicly accessible historical statistics for third-party channels, we retrieve historical observations from the Internet Archive's Wayback Machine. Specifically, we identify archived snapshots of the corresponding YouTube watch pages and extract the cumulative view counts displayed at the time of capture.

\paragraph{Observation Window.}
Directly comparing current cumulative view counts would introduce substantial differences in exposure time, as the evaluated songs were released in different years. We therefore focus on early YouTube viewership within a fixed post-release observation window.

Our primary analysis uses archived view counts observed within two days of the seventh day after each video's publication. For each video, we select the available snapshot closest to day seven and retain its original view count without interpolation or extrapolation. This yields usable historical observations for 64 of the 76 music videos: 19 from AESPA, 19 from IVE, and 26 from TWICE. The actual capture ages range from 5.2 to 8.8 days after publication.

We additionally consider two alternative settings for robustness analysis: a stricter seven-day window ($\pm 1$ day) and a 30-day window ($\pm 3$ days), covering 51 and 46 music videos, respectively. Videos without an archived observation within the corresponding window are excluded from that analysis.

\paragraph{Evaluation Protocol.}
We use the existing visual consensus rankings obtained from the 28 VLM configurations, without performing any additional model inference. For each artist group, we restrict the comparison to songs with available historical viewership data and rank these songs by their observed view counts, with higher view counts indicating greater popularity within the observation window.

We then compute Spearman's rank correlation between the model-derived visual consensus ranking and the corresponding viewership ranking. Correlations are calculated independently for each artist group rather than pooling songs across groups with different audience sizes. We report two-sided $p$-values and additionally verify the results using permutation tests with 100,000 random shuffles.

\paragraph{Robustness Analysis.}
Table~\ref{tab:market_alignment_robustness} reports the results under different observation windows. All correlations remain positive, with AESPA consistently showing the strongest association. The stricter seven-day window produces results similar to the primary analysis, while the 30-day analysis preserves the same overall pattern despite its lower data coverage.

\begin{table}[t]
\centering
\small
\setlength{\tabcolsep}{4pt}
\renewcommand{\arraystretch}{1.12}
\begin{tabular}{lccc}
\toprule
\textbf{Setting} & \textbf{AESPA} & \textbf{IVE} & \textbf{TWICE} \\
\midrule
Day 7 ($\pm 2$ days)
& 0.661 (19) & 0.363 (19) & 0.406 (26) \\
Day 7 ($\pm 1$ day)
& 0.704 (15) & 0.354 (14) & 0.423 (22) \\
Day 30 ($\pm 3$ days)
& 0.782 (11) & 0.423 (13) & 0.367 (22) \\
\bottomrule
\end{tabular}

\caption{
Robustness analysis of the association between VLM consensus rankings and historical YouTube viewership across different observation windows. Each cell reports Spearman's $\rho$, with the number of available music videos indicated in parentheses.
}
\label{tab:market_alignment_robustness}
\end{table}

We further conduct an exploratory analysis controlling for publication time. The resulting partial Spearman correlations are 0.60, 0.32, and 0.55 for AESPA, IVE, and TWICE, respectively, compared with the original correlations of 0.66, 0.36, and 0.41. These results suggest that the observed associations are not solely explained by the chronological ordering of releases.

\section{AI Assistance Disclosure}
ChatGPT was used solely for language polishing. All research design, experiments, data collection, analysis, and conclusions were independently completed by the authors.

\begin{table*}[!t]
\centering
\scriptsize
\setlength{\tabcolsep}{4pt}
\renewcommand{\arraystretch}{1.12}
\begin{tabularx}{\textwidth}{p{2.2cm} X}
\toprule
\textbf{Group} & \textbf{Songs} \\
\midrule
\textbf{AESPA} &
Lemonade; Switchblade (Feat. Ty Dolla \$ign); Supernova; Kiss N Tell; WDA (Whole Different Animal); Whiplash; Dirty Work; Drama; Live My Life; Armageddon; Rich Man; Spicy; Hot Mess; Better Things; Welcome To My World (Feat. Nævis); Life's Too Short (English Ver.); Savage; Next Level; Dreams Come True; Girls; Black Mamba; Forever; Hold On Tight; Count On Me; 'Til We Die; Aenergy; Yeppi Yeppi; Can't Help Myself; Just Another Girl; Prologue. \\
\midrule
\textbf{IVE} &
Eleven; Love Dive; After Like; Kitsch; I Am; Wave; I Want; Either Way; Off The Record; Baddie; All Night (Feat. Saweetie); Heya; Accendio; Summer Festa; Crush; Rebel Heart; Attitude; XOXZ; Bang Bang; Blackhole; Supernova Love; Lucid Dream; Be Alright; Take It; Royal; Blue Blood; Holy Moly; OTT; Payback; Ice Queen. \\
\midrule
\textbf{LE SSERAFIM} &
Hot; Crazy; Made My Night; Fearless; Perfect Night; Unforgiven; Smart; Antifragile; Boompala; Impurities; Celebration; Different; Easy; Come Over; Spaghetti; Eve, Psyche \& The Bluebeard's Wife; Kawaii; Sour Grapes; Blue Flame; The Great Mermaid; No Celestial; Good Parts (When The Quality Is Bad But I Am); Fire In The Belly; 1-800-Hot-N-Fun; Pierrot; Crazier; Ash; So Cynical (Badum); Iffy Iffy; Trust Exercise. \\
\midrule
\textbf{NEWJEANS} &
Attention; Hype Boy; Hurt; Cookie; Ditto; OMG; Zero; New Jeans; Super Shy; Cool With You (Side A); ETA; Bubble Gum; How Sweet; Right Now; Supernatural; ASAP; Get Up; Gods; Beautiful Restriction; Our Night Is More Beautiful Than Your Day. \\
\midrule
\textbf{I-DLE} &
Nxde; Tomboy; Mono; I Want That; Queencard; Oh My God; Gimme Dat Love; Good Thing; Girlfriend; Crow; Lion; Klaxon; Super Lady; Hwaa; Wife; Latata; Senorita; Uh-Oh; Hann (Alone); Allergy; Dumdi Dumdi; Revenge; I Do; Last Dance; Blow Your Mind; My Bag; Fate; I'm The Trend; Bloom; Nightmare. \\
\midrule
\textbf{TWICE} &
This Is For; Strategy; What Is Love?; The Feels; Fancy; Feel Special; Talk That Talk; I Got You; Set Me Free; One Spark; I Can't Stop Me; Alcohol-Free; Me+You; More \& More; Scientist; TT; Knock Knock; Yes Or Yes; Dance The Night Away; Likey; Cheer Up; Like Ooh-Ahh; Heart Shaker; Signal; Doughnut; Perfect World; Kura Kura; Celebrate; Hare Hare; Moonlight Sunrise. \\
\midrule
\textbf{BLACKPINK} &
As If It's Your Last; How You Like That; Playing With Fire; Go; Pink Venom; Lovesick Girls; Jump; Kill This Love; Shut Down; Boombayah; Ddu-Du Ddu-Du; Ice Cream; Whistle; Ready For Love; Stay; Typa Girl; Forever Young; Really; See U Later; Don't Know What To Do; Kick It; Hope Not; Pretty Savage; Bet You Wanna; Crazy Over You; Love To Hate Me; You Never Know; Yeah Yeah Yeah; The Happiest Girl; Tally. \\
\midrule
\textbf{RED VELVET} &
Russian Roulette; Cosmic; Feel My Rhythm; Surfin Boy; Chill Kill; Psycho; Peek-A-Boo; Dumb Dumb; Zimzalabim; Red Flavor; Bad Boy; RBB (Really Bad Boy); Queendom; Umpah Umpah; Ice Cream Cake; Happiness; Cookie Jar; Automatic; One Of These Nights; Rookie; Power Up; Would U; Rebirth; I Just; Sappy; Wildside; Be Natural; Wish Tree; Look; Kingdom Come. \\
\midrule
\textbf{ITZY} &
Wannabe; Tunnel Vision; Loco; Bet On Me; ICY; Motto; In The Morning; Dalla Dalla; Girls Will Be Girls; Cheshire; Untouchable; Not Shy; Born To Be; Imaginary Friend; Cake; Boys Like You; Gold; None Of My Business; Sneakers; Swipe; Mr. Vampire; Voltage; Blah Blah Blah; Ringo; Algorhythm; Rock \& Roll; Shoot!; That's A No No; \#Twenty; Sorry Not Sorry. \\
\midrule
\textbf{NMIXX} &
Blue Valentine; Heavy Serenade; Know About Me; Crescendo; Dice; Love Me Like This; Dash; Party O'clock; Tic Tic (Feat. Pabllo Vittar); See That?; Soñar (Breaker); Roller Coaster; O.O; Young, Dumb, Stupid; Funky Glitter Christmas; Feifei; Run For Roses; Cool (Your Rainbow); Passionfruit; High Horse; Spinnin' On It; Papillon; Boom; Love Is Lonely; XOXO; My Gosh; Paxxword; Beat Beat; Red Light Sign, But We Go; Break The Wall. \\
\midrule
\textbf{BABYMONSTER} &
Batter Up; Stuck In The Middle; Sheesh; Like That; Dream; Forever; Clik Clak; Drip; Love, Maybe; Really Like You; Billionaire; Love In My Heart; Woke Up In Tokyo; Ghost; Hot Sauce; We Go Up; Psycho; Supa Dupa Luv; Wild; Moon; Choom; I Like It; Locked In; Sugar Honey Ice Tea. \\
\midrule
\textbf{MEOVV} &
Meow; Toxic; Body; Hands Up; Drop Top; Lit Right Now; Me Me Me; Burning Up; Hit 'Em; DDI Ro Ri; In My Hands; Favorite Song; Revenge. \\
\midrule
\textbf{ILLIT} &
My World; Magnetic; Midnight Fiction; Lucky Girl Syndrome; I'll Like You; Cherish (My Love); IYKYK; Pimple; Tick-Tack; Almond Chocolate; I Got Your Back; Little Monster; Billyeoon Goyangi (Do The Dance); Jellyous; Oops!; Bamsopoong; Topping; Toki Yo Tomare; Secret Quest; Not Cute Anymore; Not Me; All For You; Sunday Morning; GRWM (Get Ready With Me); It's Me; Paw, Paw!; Mamihlapinatapai; Love, Older You; Swingin' Magic; Bubee. \\
\midrule
\textbf{HEARTS2HEARTS} &
The Chase; Butterflies; Style; Pretty Please; Focus; Apple Pie; Flutter; Blue Moon; Rude!; Lemon Tang; 15-Love; Baby Steps; Heart Emoji (\$\textbackslash\{\}heartsuit\$); Secret Recipe; Iconic Heart; Moonride. \\
\bottomrule
\end{tabularx}
\caption{Complete list of audio tracks included in the audio evaluation (373 items across 14 K-pop groups).}
\label{tab:full_audio_list}
\end{table*}

\begin{table*}[!t]
\centering
\scriptsize
\setlength{\tabcolsep}{4pt}
\renewcommand{\arraystretch}{1.12}
\begin{tabularx}{\textwidth}{p{2.2cm} X}
\toprule
\textbf{Group} & \textbf{Songs} \\
\midrule
\textbf{AESPA} &
Lemonade; Switchblade (Feat. Ty Dolla \$ign); Supernova; Kiss N Tell; WDA (Whole Different Animal); Whiplash; Dirty Work; Drama; Live My Life; Armageddon; Rich Man; Spicy; Hot Mess; Better Things; Welcome To My World (Feat. Nævis); Life's Too Short (English Ver.); Savage; Next Level; Dreams Come True; Girls; Black Mamba; Forever; Hold On Tight. \\
\midrule
\textbf{IVE} &
Eleven; Love Dive; After Like; Kitsch; I Am; Wave; I Want; Either Way; Off The Record; Baddie; All Night (Feat. Saweetie); Heya; Accendio; Summer Festa; Crush; Rebel Heart; Attitude; XOXZ; Bang Bang; Blackhole; Supernova Love; Lucid Dream; Be Alright. \\
\midrule
\textbf{LE SSERAFIM} &
Hot; Crazy; Made My Night; Fearless; Perfect Night; Unforgiven; Smart; Antifragile; Boompala; Impurities; Celebration; Different; Easy; Come Over; Spaghetti; Eve, Psyche \& The Bluebeard's Wife; Kawaii. \\
\midrule
\textbf{NEWJEANS} &
Attention; Hype Boy; Hurt; Cookie; Ditto; OMG; Zero; New Jeans; Super Shy; Cool With You (Side A); ETA; Bubble Gum; How Sweet; Right Now; Supernatural; ASAP. \\
\midrule
\textbf{I-DLE} &
Nxde; Tomboy; Mono; I Want That; Queencard; Oh My God; Gimme Dat Love; Good Thing; Girlfriend; Crow; Lion; Klaxon; Super Lady; Hwaa; Wife; Latata; Senorita; Uh-Oh; Hann (Alone); Allergy; Dumdi Dumdi; Revenge; I Do; Last Dance; Blow Your Mind. \\
\midrule
\textbf{TWICE} &
This Is For; Strategy; What Is Love?; The Feels; Fancy; Feel Special; Talk That Talk; I Got You; Set Me Free; One Spark; I Can't Stop Me; Alcohol-Free; Me+You; More \& More; Scientist; TT; Knock Knock; Yes Or Yes; Dance The Night Away; Likey; Cheer Up; Like Ooh-Ahh; Heart Shaker; Signal; Doughnut; Perfect World; Kura Kura; Celebrate; Hare Hare; Moonlight Sunrise. \\
\midrule
\textbf{BLACKPINK} &
As If It's Your Last; How You Like That; Playing With Fire; Go; Pink Venom; Lovesick Girls; Jump; Kill This Love; Shut Down; Boombayah; Ddu-Du Ddu-Du; Ice Cream; Whistle; Ready For Love; Stay; Typa Girl; Forever Young; Really; See U Later; Don't Know What To Do; Kick It; Hope Not; Pretty Savage; Bet You Wanna; Crazy Over You; Love To Hate Me; You Never Know; Yeah Yeah Yeah; The Happiest Girl; Tally. \\
\midrule
\textbf{RED VELVET} &
Russian Roulette; Cosmic; Feel My Rhythm; Surfin Boy; Chill Kill; Psycho; Peek-A-Boo; Dumb Dumb; Zimzalabim; Red Flavor; Bad Boy; RBB (Really Bad Boy); Queendom; Umpah Umpah; Ice Cream Cake; Happiness; Cookie Jar; Automatic; One Of These Nights; Rookie; Power Up; Would U; Rebirth; I Just; Sappy; Wildside; Be Natural; Wish Tree. \\
\midrule
\textbf{ITZY} &
Wannabe; Tunnel Vision; Loco; Bet On Me; ICY; Motto; In The Morning; Dalla Dalla; Girls Will Be Girls; Cheshire; Untouchable; Not Shy; Born To Be; Imaginary Friend; Cake; Boys Like You; Gold; None Of My Business; Sneakers; Swipe; Mr. Vampire; Voltage; Blah Blah Blah; Ringo; Algorhythm; Rock \& Roll. \\
\midrule
\textbf{NMIXX} &
Blue Valentine; Heavy Serenade; Know About Me; Crescendo; Dice; Love Me Like This; Dash; Party O'clock; Tic Tic (Feat. Pabllo Vittar); See That?; Soñar (Breaker); Roller Coaster; O.O; Young, Dumb, Stupid; Funky Glitter Christmas; Feifei. \\
\midrule
\textbf{BABYMONSTER} &
Batter Up; Stuck In The Middle; Sheesh; Forever; Clik Clak; Drip; Really Like You; Love In My Heart; Hot Sauce; We Go Up; Psycho; Supa Dupa Luv; Moon; Choom; I Like It; Sugar Honey Ice Tea. \\
\midrule
\textbf{MEOVV} &
Meow; Toxic; Hands Up; Drop Top; Burning Up; DDI Ro Ri; In My Hands. \\
\midrule
\textbf{ILLIT} &
Jellyous; Not Cute Anymore; It's Me; Tick-Tack; Cherish (My Love); Billyeoon Goyangi (Do The Dance); Lucky Girl Syndrome; Magnetic; Sunday Morning; I Got Your Back; Not Me; Toki Yo Tomare. \\
\midrule
\textbf{HEARTS2HEARTS} &
The Chase; Butterflies; Pretty Please; Style; Focus; Rude!; Iconic Heart; Lemon Tang; Moonride. \\
\bottomrule
\end{tabularx}
\caption{Complete list of music videos included in the mv evaluation (278 items across 14 K-pop groups).}
\label{tab:full_mv_list}
\end{table*}

\begin{sidewaystable*}[t]
\centering
\scriptsize
\setlength{\tabcolsep}{2.0pt}

\caption{
Pairwise ranking agreement between VLMs in the music-video setting. Each entry reports the Spearman rank correlation coefficient averaged across 14 artist groups. Higher values indicate stronger agreement between model rankings.
}

\label{tab:video_spearman_full}
\resizebox{\textwidth}{!}{
\begin{tabular}{lcccccccccccccccccccccccccccc}
\toprule
\textbf{Model}
& \textbf{G-N} & \textbf{G-T}
& \textbf{I1-N} & \textbf{I1-T}
& \textbf{I2-N} & \textbf{I2-T}
& \textbf{I4-N} & \textbf{I4-T}
& \textbf{I8-N} & \textbf{I8-T}
& \textbf{I14-N} & \textbf{I14-T}
& \textbf{I38-N} & \textbf{I38-T}
& \textbf{M-N} & \textbf{M-T}
& \textbf{Q2-N} & \textbf{Q2-T}
& \textbf{Q4-N} & \textbf{Q4-T}
& \textbf{Q9-N} & \textbf{Q9-T}
& \textbf{Q27-N} & \textbf{Q27-T}
& \textbf{V2} & \textbf{V4} & \textbf{V8} & \textbf{V32} \\
\midrule
G-N & -- & 0.895 & 0.326 & 0.307 & 0.476 & 0.474 & 0.648 & 0.629 & 0.582 & 0.709 & 0.655 & 0.721 & 0.662 & 0.722 & 0.600 & 0.623 & 0.423 & 0.637 & 0.630 & 0.650 & 0.507 & 0.585 & 0.554 & 0.506 & 0.595 & 0.541 & 0.615 & 0.641 \\
G-T &  & -- & 0.351 & 0.335 & 0.478 & 0.523 & 0.708 & 0.704 & 0.587 & 0.751 & 0.674 & 0.711 & 0.678 & 0.791 & 0.618 & 0.656 & 0.469 & 0.721 & 0.626 & 0.684 & 0.511 & 0.610 & 0.581 & 0.577 & 0.599 & 0.616 & 0.661 & 0.599 \\
I1-N &  &  & -- & 0.604 & 0.565 & 0.591 & 0.538 & 0.559 & 0.530 & 0.507 & 0.413 & 0.332 & 0.523 & 0.389 & 0.189 & 0.244 & 0.438 & 0.258 & 0.110 & 0.084 & 0.138 & 0.051 & 0.010 & -0.019 & 0.415 & 0.357 & 0.254 & 0.249 \\
I1-T &  &  &  & -- & 0.654 & 0.825 & 0.579 & 0.681 & 0.731 & 0.645 & 0.593 & 0.453 & 0.617 & 0.464 & 0.306 & 0.318 & 0.620 & 0.353 & 0.221 & 0.181 & 0.250 & 0.101 & 0.106 & -0.033 & 0.529 & 0.534 & 0.331 & 0.279 \\
I2-N &  &  &  &  & -- & 0.832 & 0.706 & 0.721 & 0.698 & 0.688 & 0.570 & 0.510 & 0.643 & 0.562 & 0.382 & 0.436 & 0.477 & 0.355 & 0.308 & 0.300 & 0.210 & 0.238 & 0.208 & 0.091 & 0.566 & 0.419 & 0.333 & 0.320 \\
I2-T &  &  &  &  &  & -- & 0.751 & 0.818 & 0.828 & 0.795 & 0.678 & 0.578 & 0.720 & 0.638 & 0.451 & 0.489 & 0.587 & 0.441 & 0.332 & 0.325 & 0.238 & 0.218 & 0.219 & 0.089 & 0.646 & 0.557 & 0.412 & 0.348 \\
I4-N &  &  &  &  &  &  & -- & 0.920 & 0.736 & 0.832 & 0.717 & 0.714 & 0.714 & 0.714 & 0.505 & 0.544 & 0.568 & 0.603 & 0.512 & 0.505 & 0.401 & 0.447 & 0.380 & 0.330 & 0.662 & 0.618 & 0.529 & 0.508 \\
I4-T &  &  &  &  &  &  &  & -- & 0.807 & 0.881 & 0.757 & 0.735 & 0.749 & 0.749 & 0.509 & 0.556 & 0.634 & 0.590 & 0.499 & 0.498 & 0.409 & 0.430 & 0.351 & 0.303 & 0.649 & 0.662 & 0.545 & 0.518 \\
I8-N &  &  &  &  &  &  &  &  & -- & 0.875 & 0.852 & 0.783 & 0.847 & 0.737 & 0.532 & 0.544 & 0.686 & 0.525 & 0.456 & 0.427 & 0.449 & 0.327 & 0.312 & 0.184 & 0.682 & 0.645 & 0.560 & 0.530 \\
I8-T &  &  &  &  &  &  &  &  &  & -- & 0.866 & 0.841 & 0.852 & 0.860 & 0.623 & 0.666 & 0.649 & 0.659 & 0.610 & 0.594 & 0.499 & 0.481 & 0.458 & 0.353 & 0.736 & 0.678 & 0.669 & 0.633 \\
I14-N &  &  &  &  &  &  &  &  &  &  & -- & 0.931 & 0.829 & 0.804 & 0.619 & 0.613 & 0.674 & 0.660 & 0.609 & 0.574 & 0.557 & 0.466 & 0.480 & 0.348 & 0.684 & 0.691 & 0.672 & 0.663 \\
I14-T &  &  &  &  &  &  &  &  &  &  &  & -- & 0.798 & 0.816 & 0.618 & 0.618 & 0.632 & 0.660 & 0.662 & 0.607 & 0.611 & 0.538 & 0.551 & 0.427 & 0.664 & 0.652 & 0.669 & 0.711 \\
I38-N &  &  &  &  &  &  &  &  &  &  &  &  & -- & 0.816 & 0.581 & 0.609 & 0.642 & 0.585 & 0.554 & 0.503 & 0.516 & 0.398 & 0.458 & 0.342 & 0.711 & 0.630 & 0.619 & 0.616 \\
I38-T &  &  &  &  &  &  &  &  &  &  &  &  &  & -- & 0.694 & 0.736 & 0.549 & 0.673 & 0.692 & 0.707 & 0.559 & 0.597 & 0.586 & 0.520 & 0.633 & 0.633 & 0.729 & 0.689 \\
M-N &  &  &  &  &  &  &  &  &  &  &  &  &  &  & -- & \textbf{0.932} & 0.454 & 0.722 & 0.735 & 0.787 & 0.626 & 0.740 & 0.730 & 0.658 & 0.489 & 0.574 & 0.788 & 0.617 \\
M-T &  &  &  &  &  &  &  &  &  &  &  &  &  &  &  & -- & 0.439 & 0.726 & 0.705 & 0.806 & 0.573 & 0.734 & 0.695 & 0.663 & 0.499 & 0.583 & 0.783 & 0.600 \\
Q2-N &  &  &  &  &  &  &  &  &  &  &  &  &  &  &  &  & -- & 0.597 & 0.419 & 0.406 & 0.490 & 0.346 & 0.318 & 0.249 & 0.501 & 0.602 & 0.506 & 0.445 \\
Q2-T &  &  &  &  &  &  &  &  &  &  &  &  &  &  &  &  &  & -- & 0.708 & 0.813 & 0.596 & 0.747 & 0.666 & 0.677 & 0.540 & 0.713 & 0.765 & 0.623 \\
Q4-N &  &  &  &  &  &  &  &  &  &  &  &  &  &  &  &  &  &  & -- & 0.846 & 0.735 & 0.771 & 0.767 & 0.678 & 0.548 & 0.529 & 0.757 & 0.703 \\
Q4-T &  &  &  &  &  &  &  &  &  &  &  &  &  &  &  &  &  &  &  & -- & 0.671 & 0.914 & 0.799 & 0.811 & 0.461 & 0.601 & 0.787 & 0.650 \\
Q9-N &  &  &  &  &  &  &  &  &  &  &  &  &  &  &  &  &  &  &  &  & -- & 0.707 & 0.712 & 0.621 & 0.441 & 0.578 & 0.727 & 0.688 \\
Q9-T &  &  &  &  &  &  &  &  &  &  &  &  &  &  &  &  &  &  &  &  &  & -- & 0.803 & 0.843 & 0.367 & 0.560 & 0.731 & 0.642 \\
Q27-N &  &  &  &  &  &  &  &  &  &  &  &  &  &  &  &  &  &  &  &  &  &  & -- & 0.851 & 0.452 & 0.496 & 0.713 & 0.682 \\
Q27-T &  &  &  &  &  &  &  &  &  &  &  &  &  &  &  &  &  &  &  &  &  &  &  & -- & 0.281 & 0.489 & 0.667 & 0.567 \\
V2 &  &  &  &  &  &  &  &  &  &  &  &  &  &  &  &  &  &  &  &  &  &  &  &  & -- & 0.558 & 0.573 & 0.629 \\
V4 &  &  &  &  &  &  &  &  &  &  &  &  &  &  &  &  &  &  &  &  &  &  &  &  &  & -- & 0.723 & 0.595 \\
V8 &  &  &  &  &  &  &  &  &  &  &  &  &  &  &  &  &  &  &  &  &  &  &  &  &  &  & -- & 0.750 \\
V32 &  &  &  &  &  &  &  &  &  &  &  &  &  &  &  &  &  &  &  &  &  &  &  &  &  &  &  & -- \\
\bottomrule
\end{tabular}
}
\end{sidewaystable*}

\end{document}